\documentclass[sigconf]{acmart}
\AtBeginDocument{%
  }
\usepackage{multirow}

\definecolor{hk}{RGB}{0,102,204}
\definecolor{dtcolor}{RGB}{3, 219, 252}

\usepackage{xspace}

\setcopyright{cc}
\setcctype{by}
\copyrightyear{2026}
\acmYear{2026}
\acmDOI{10.1145/3767308.3835973}
\acmConference[MM '26]{Proceedings of the 34th ACM International Conference on Multimedia}{November 10--14, 2026}{Rio de Janeiro, Brazil}
\acmBooktitle{Proceedings of the 34th ACM International Conference on Multimedia (MM '26), November 10--14, 2026, Rio de Janeiro, Brazil}

\acmISBN{979-8-4007-2213-4/2026/11}

\acmSubmissionID{5893}

\begin{document}

\title[Kinematics-Centric Continuous Sign Language Retrieval with Gloss-Guided Boundary-Aware Alignment]{Kinematics-Centric Continuous Sign Language Retrieval with Gloss-Guided Boundary-Aware Alignment}

\author{Chang Liu}
\email{chang.liu-2@unitn.it}
\affiliation{%
  \institution{University of Trento}
  \city{Trento}
  \country{Italy}
}

\author{Ke Han}
\authornote{Corresponding author.}
\email{ke.han.aca@gmail.com}
\affiliation{%
  \institution{University of Trento}
  \city{Trento}
  \country{Italy}
}

\author{Davide Talon}
\email{dtalon@fbk.eu}
\affiliation{%
  \institution{Fondazione Bruno Kessler}
  \city{Trento}
  \country{Italy}
}

\author{Elisa Ricci}
\affiliation{%
  \institution{University of Trento}
  \city{Trento}
  \country{Italy}
}
\affiliation{%
  \institution{Fondazione Bruno Kessler}
  \city{Trento}
  \country{Italy}
}
\email{e.ricci@unitn.it}

\author{Nicu Sebe}
\email{niculae.sebe@unitn.it}
\affiliation{%
  \institution{University of Trento}
  \city{Trento}
  \country{Italy}
}


\renewcommand{\shortauthors}{Liu et al.}

\begin{abstract}
Sign language–text alignment remains a fundamental challenge for text-driven sign language understanding. Existing methods predominantly rely on appearance-heavy RGB representations, which entangle motion semantics with visual variations and lead to ambiguous motion–language grounding.
In this paper, we reformulate sign language–text alignment in a structured kinematic space and propose a kinematics-centric framework that adopts 3D motion parameterized by SMPL-eXpressive (SMPL-X) as the primary representation. By explicitly modeling the kinematic dynamics of signing in a unified motion space, our approach reduces reliance on appearance signals and yields more semantically consistent representations.
To capture the compositional nature of sign language, we introduce a gloss-guided local alignment mechanism that leverages gloss temporal spans as automatically estimated training-time supervision to decompose continuous motion into coherent segments and establish fine-grained motion–text correspondences, thereby reducing ambiguity in localizing word-level semantics in continuous signing. Furthermore, we develop a visual distillation strategy, where RGB signals serve as privileged supervision during training to provide complementary contextual cues, and retrieval is performed using only SMPL-X motion and text representations.
Extensive experiments on standard benchmarks demonstrate that our method achieves state-of-the-art bidirectional retrieval performance on CSL-Daily and competitive results on PHOENIX-2014T. These results highlight the effectiveness of kinematic representations and explicit local grounding for sign language–text alignment.
\end{abstract}



\maketitle
\section{Introduction}
\begin{figure*}[h]
  \centering
  \includegraphics[width=0.8\linewidth]{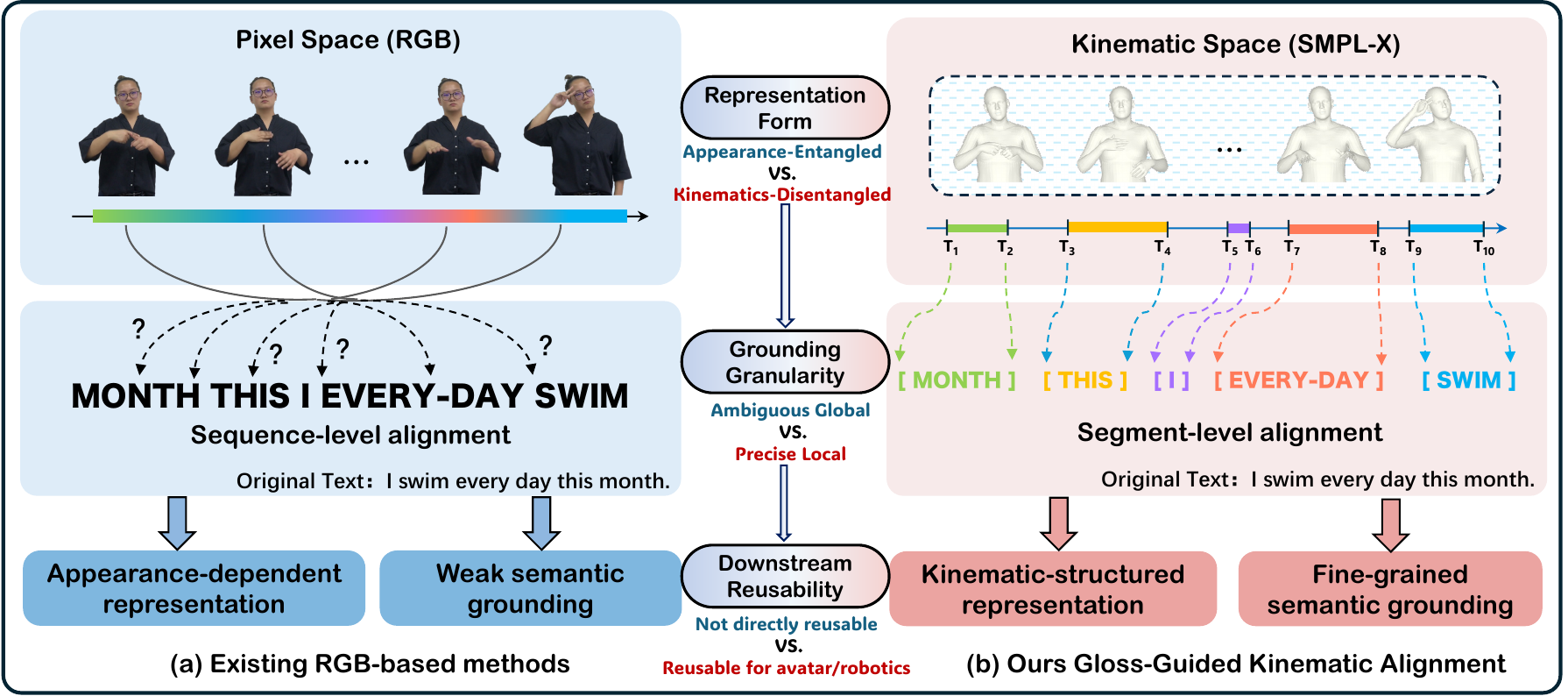}
    \caption{Comparison of RGB-based sign language-text alignment and our gloss-guided kinematic alignment.}  
    \label{fig:first}
\end{figure*}
With the advancement of multimodal artificial intelligence, sign language–text applications have advanced rapidly, offering significant benefits to individuals with hearing and speech impairments.
Sign language–text alignment is fundamental to enabling text-driven sign language applications, such as sign language retrieval, generation, and animation. To this end, existing methods typically learn a shared embedding space between sign sequences and natural language through contrastive learning or cross-modal matching objectives \cite{cheng2023cico, jiang2024seds}, progressively improving cross-modal alignment.

Sign language is inherently expressed through articulated motion in 3D space, involving coordinated movements of the hands, body, and face, and is largely independent of appearance cues. However, most existing methods focus on aligning text with visual representations extracted from raw RGB videos, making them heavily reliant on appearance information \cite{cheng2023cico,jiang2024seds,wu2024uncertainty,duarte2022sign}.
As a result, RGB-based representations tend to entangle semantic motion with signer-specific appearance and viewpoints, and are also limited in capturing the underlying 3D motion dynamics, potentially leading to ambiguous motion–language grounding.
Furthermore, being rooted in the visual modality rather than structured motion representations, they are inherently incompatible with motion-centric downstream tasks, such as Gaussian-splatting-based avatar reconstruction, kinematic sign language generation, and robot manipulation \cite{kocabas2024hugs, ivashechkin2025signsplat, yu2024signavatars, xu2025dexcanvas}.

To address these limitations, we reformulate sign language–text alignment in a structured motion space, using SMPL-X~\cite{pavlakos2019expressive} as the primary kinematic representation rather than a visual space.
As a unified parametric 3D human model of the body and hands, SMPL-X provides a structured motion representation that better reflects the articulatory nature of sign language.
Compared with RGB-based representations, it shifts the focus from appearance-driven visual signals to explicit human kinematics in a canonical and geometrically interpretable 3D form, while being more compatible with motion-centric downstream tasks and more privacy-friendly than raw RGB videos \cite{moryossef2021evaluating, fish2025geo, rust2024towards}.
In this sense, SMPL-X is not merely an input modality, but a structured semantic representation compatible with motion-centric deployment for sign language alignment.

However, precise alignment between text and SMPL-X-based motion remains challenging. 
Unlike RGB videos, which provide rich contextual and appearance cues, SMPL-X encodes signing in a structured yet abstract parametric space. In this space, semantics arise from coordinated movements of multiple articulators, including the hands, body, and face, rather than from isolated body parts, making cross-articulator dependencies difficult to model. 
Moreover, prior methods predominantly rely on global alignment between sentence-level text and sequence-level sign motion~\cite{cheng2023cico,wu2024uncertainty}, overlooking fine-grained correspondences between localized lexical semantics and their associated motion segments. Consequently, the learned representations remain entangled at the sequence level and are less robust for local semantic grounding, as illustrated in Fig.~\ref{fig:first}.

To address these challenges, we propose a Gloss-Guided Kinematic Sign Language-Text Alignment framework. 
Instead of treating alignment as a purely global sequence-level matching problem, we reformulate it as a structured segment-level semantic alignment task in kinematic space. Specifically, we adopt SMPL-X as the primary motion representation to capture the articulatory dynamics of sign language beyond appearance-dependent visual features. 
We further leverage gloss annotations as automatically estimated training-time supervision to anchor lexical semantics in time. Here, gloss is symbolic label used to annotate localized sign units in continuous signing.  Importantly, the order of glosses does not necessarily match the word order of the corresponding spoken-language sentence \cite{feng2022language, stoll2018sign, sandler2006sign}, so gloss supervision serves as a temporal semantic anchor rather than a strict word-by-word alignment signal. Based on gloss spans, we decompose continuous motion into semantically coherent segments and align each with relevant textual content. 
A segment-level contrastive objective further improves local consistency and reduces cross-segment interference.

To compensate for this abstraction, we introduce an auxiliary visual distillation strategy.
Although RGB signals may introduce appearance variations when used as primary representations, they still provide contextual cues during training. Therefore, we treat RGB as privileged supervision only in training: an auxiliary RGB stream provides dense spatiotemporal context, and its knowledge is distilled into the SMPL-X motion branch. During retrieval, the model operates on SMPL-X motion and text representations while preserving the benefits of visual supervision during training.


We evaluate the proposed framework on sign language retrieval benchmarks. On CSL-Daily, our method achieves state-of-the-art bidirectional retrieval performance. Our method also remains competitive on the more challenging PHOENIX-2014T benchmark, highlighting both the promise of kinematics-centric retrieval and the importance of motion quality in difficult visual conditions.

The main contributions of this work are summarized as follows:
\begin{enumerate}
    \item We propose a kinematics-centric framework that aligns sign language with text in 3D kinematic space, replacing appearance-dependent matching with structured semantic alignment. Our approach performs bidirectional retrieval in the SMPL-X--text embedding space and improves compatibility with downstream applications.
    
    \item We develop a gloss-guided local semantic alignment mechanism that leverages gloss temporal spans as automatically estimated training-time supervision to decompose continuous signing motion into semantically coherent segments, enabling fine-grained motion-text correspondence and improving local semantic grounding.
    
    \item We introduce a visual distillation strategy, where RGB serves as privileged supervision during training to provide complementary contextual cues, with its knowledge distilled into the motion representation used for retrieval.
    
    \item Extensive experiments demonstrate that our method achieves state-of-the-art bidirectional retrieval performance on CSL-Daily and competitive results on PHOENIX-2014T, highlighting the effectiveness of structured kinematic representations for sign language-text alignment.
\end{enumerate}

\section{Related Work}

\subsection{Visual Representations for Sign Language}

Visual representation is a central design choice in sign language understanding, including recognition, translation, and retrieval. Raw RGB video remains the dominant input modality in these tasks~\cite{carreira2017quo, cheng2023cico, jiang2024seds, zheng2023cvt}. Benefiting from strong spatiotemporal backbones and large-scale pre-training, RGB-based models can capture rich appearance, contextual, and motion cues, and have achieved strong empirical results across multiple tasks. Nevertheless, RGB is not an ideal representation of sign language structure. Sign semantics are primarily carried by articulated human motion, including hand movement, body pose, and facial expressions. In raw RGB frames, these linguistically relevant cues are inevitably entangled with background, illumination, clothing patterns, and signer-specific appearance~\cite{duarte2022sign, cheng2023cico, jiang2024seds, wu2024uncertainty}. Such visual entanglement can bias models toward superficial context or identity cues and weaken signer-invariant semantic modeling~\cite{wu2024uncertainty, wong2025signrep}. Moreover, RGB remains a high-dimensional pixel representation that is difficult to reuse once retrieval is completed: although it can support semantic matching, it does not naturally provide an editable or retargetable motion representation for downstream applications.

To reduce appearance dependency, several studies have explored human structure, such as 2D skeleton keypoints, in sign language understanding~\cite{lee2023human, duarte2022sign, camgoz2020sign}. Compared with RGB, 2D pose provides a compact and signer-invariant description of motion and removes much of the irrelevant texture information. This line of work suggests that structured human motion is beneficial for sign-related tasks. However, 2D representations remain inherently incomplete for continuous signing: they lack depth information, are sensitive to self-occlusion and overlapping fingers, and often fail to capture subtle hand articulation and non-manual signals such as head movement, eyebrow motion, and facial expression~\cite{kratimenos2021independent, hu2021signbert, jiao2023cosign}.

More recently, parametric 3D human models, especially SMPL-X~\cite{pavlakos2019expressive}, have emerged as a unified representation of body, hands, and face. By organizing these articulators within a coherent topology, SMPL-X provides a structured and geometrically interpretable motion space that is well aligned with the linguistic form of signing. In sign language-related research, such 3D representations have been increasingly adopted in isolated sign recognition, signing avatar reconstruction, and sign language production, where they serve as reusable intermediates for motion editing, retargeting, and generation~\cite{kratimenos2021independent, dong2024signavatar, yu2024signavatars, baltatzis2024neural, zuo2025signs}. 
Recent pose-centric systems such as SignLLM~\cite{fang2025signllm} and SignX~\cite{fang2025signx} target sign production and continuous recognition, respectively, rather than bidirectional sign--text retrieval, so their task-specific results are not directly comparable to ours.
Beyond sign language, structured motion representations have also been widely used in text-to-motion generation and motion-language modeling, demonstrating their value for grounding language in articulated human behavior~\cite{hanmotion, guo2022generating}. Nevertheless, such structured 3D motion has rarely been adopted as the primary representation for continuous sign language retrieval, where existing frameworks remain largely RGB-dominant.

\subsection{Sign Language Retrieval}

Sign language retrieval aims to learn a shared semantic space between continuous sign sequences and natural language, enabling bidirectional matching between signed content and textual queries~\cite{duarte2022sign, cheng2023cico, wu2024uncertainty, jiang2024seds}. Different from continuous sign language recognition (CSLR), which focuses on gloss or label prediction, and sign language translation (SLT), which targets sentence generation, retrieval is fundamentally a cross-modal matching problem. As a result, it is particularly sensitive to two factors: the quality of the underlying sign representation and the granularity at which motion and language are aligned.

Existing continuous sign language retrieval methods are largely inherited from the broader text-video retrieval literature~\cite{luo2021clip4clip}. Following this paradigm, most approaches encode a complete sign video into a dense visual representation and align it with a sentence-level embedding derived from natural language~\cite{duarte2022sign}. This formulation has demonstrated promising performance and established retrieval as a practical task in sign language understanding. However, it still treats signing primarily as a generic visual stream, where semantics are inferred from appearance and motion cues.

Despite the success of global sequence-to-sentence matching, such a formulation does not fully capture the internal compositional structure of continuous signing. This motivates more fine-grained sign-text alignment beyond holistic sequence-level matching.

\subsection{Gloss Supervision for Sign-Text Alignment}
Cross-modal alignment is a central challenge in continuous sign language retrieval. Most existing methods align sign sequences with a sentence-level text embedding through global contrastive learning~\cite{duarte2022sign, cheng2023cico, wu2024uncertainty, jiang2024seds}. While effective for coarse semantic matching, such sequence-level alignment is insufficient for continuous signing, where meaning is distributed across localized lexical units and grammatical cues, making motion-text grounding difficult.

To improve semantic sensitivity, recent work has begun to incorporate gloss supervision into the retrieval process. Existing approaches explore implicit sign-to-word correspondence learning, gloss-aware fusion, attention-based pooling, optimal-transport-based matching, and soft token-frame interactions~\cite{yin2023gloss, zhou2021spatial, min2021visual, jiao2024visual, chen2024signvtcl}. These methods move beyond purely holistic matching and partially enhance local motion-language correspondence. However, in most cases gloss information is still used in an implicit or soft manner, rather than as explicit boundary-grounded supervision.

This limitation becomes especially significant in continuous signing, where strong co-articulation causes adjacent glosses to blend together and makes semantic boundaries inherently ambiguous. Under such conditions, soft local alignment can easily mix neighboring units and transfer boundary uncertainty into semantic interference, leading to imprecise local grounding~\cite{guo2023denoising, jiang2024seds}. Different from these approaches, our framework uses gloss temporal spans to construct explicit boundary-aware alignment directly in the SMPL-X motion space, enabling more precise motion-text correspondence learning while using RGB only as privileged supervision during training and SMPL-X motion for retrieval.

\section{Method}
\begin{figure*}[h]
  \centering
  \includegraphics[width=0.9\linewidth]{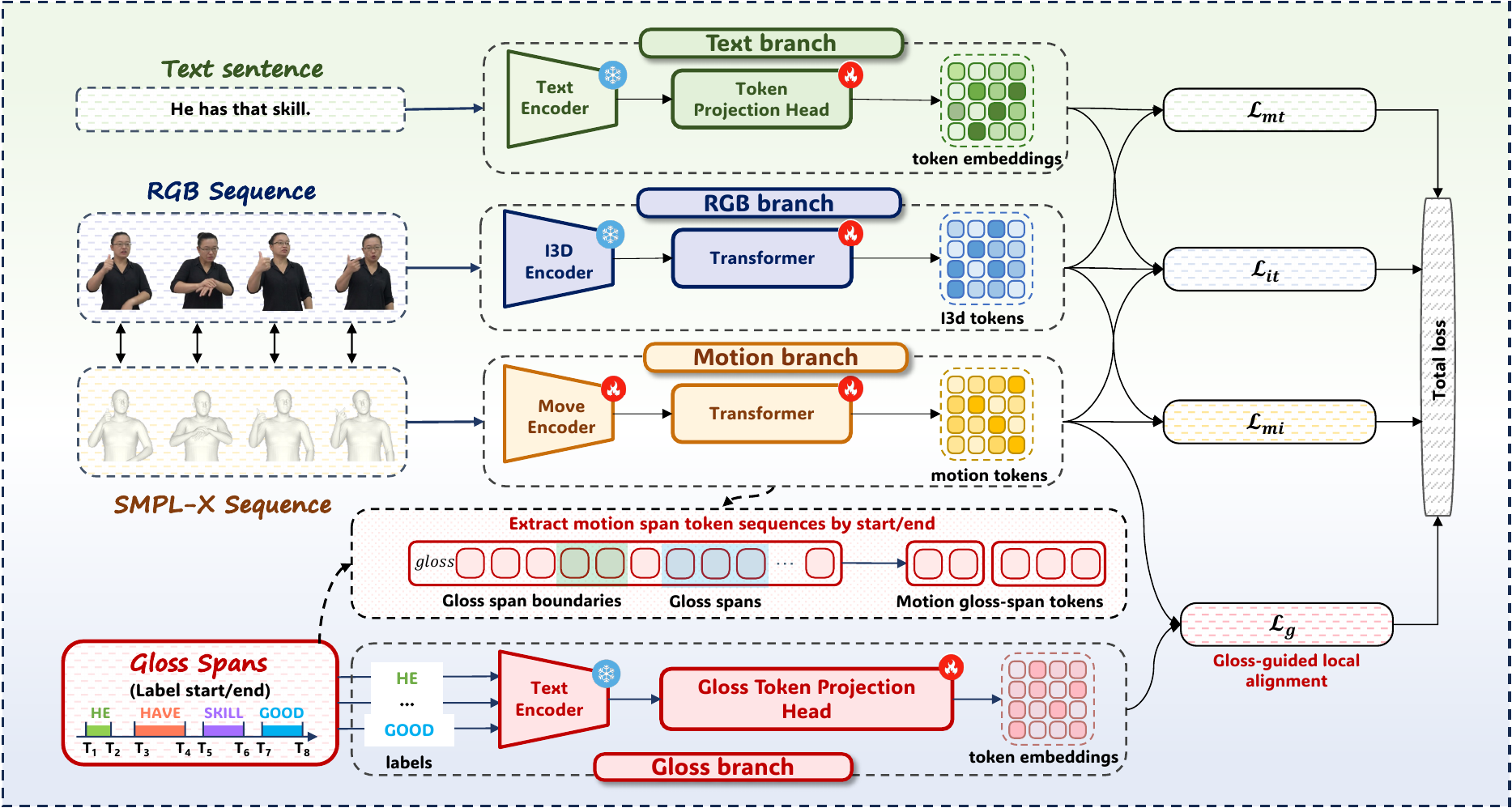}
    \caption{Overview of the proposed gloss-guided kinematic sign language--text alignment framework. During training, the model jointly exploits text, RGB, motion, and gloss supervision to learn cross-modal alignment, where gloss spans provide local motion-language grounding and RGB serves as privileged supervision. Retrieval uses only the text and motion branches.}
  \label{fig:pipeline}
\end{figure*}

This section presents our method for continuous sign language retrieval. We first define the task setting and notation, and then describe the overall framework and its main training objectives.

\textit{Preliminaries.}
We study motion--text retrieval for continuous sign language, where the goal is to learn a shared embedding space between a motion sequence and its corresponding natural-language description. At inference, given a motion sequence or text query, the model retrieves the most relevant sample from the other modality.

During training, we assume access to auxiliary supervision from RGB videos and gloss span annotations. Formally, each training sample is denoted as $(x,\mathbf{M},\mathbf{V},\mathcal{G})$, where $x$ is the input sentence, $\mathbf{M}\in\mathbb{R}^{T\times d_m}$ is the SMPL-X motion sequence, $\mathbf{V}$ is the corresponding RGB video, and
$
\mathcal{G}=\{(g_k,s_k,e_k)\}_{k=1}^{K}
$
is the set of gloss spans, with $g_k$ the $k$-th gloss label and $(s_k,e_k)$ its temporal boundaries. The auxiliary RGB and gloss annotations are only used during training, while retrieval operates solely on the motion and text representations.

\subsection{Overview}

As illustrated in Fig.~\ref{fig:pipeline}, our framework is centered on SMPL-X motion as the primary retrieval representation and is trained with auxiliary RGB and gloss supervision. The model contains four encoders for text, motion, RGB, and gloss inputs. Among them, the motion and text branches define the shared embedding space used for bidirectional retrieval, while the RGB and gloss branches are introduced during training to provide complementary supervision.

The framework is built upon three key components. First, \emph{modality representation} learning  maps each input stream into its own feature space and produces retrieval-oriented embeddings (Section~\ref{sec:method:modality-representations}). Second, \emph{gloss-guided boundary-aware local alignment} explicitly associates localized motion segments with gloss-level semantics, improving fine-grained grounding between sign motion and linguistic units (Section~\ref{sec:method:gloss-guided}). Third, \emph{auxiliary visual distillation} transfers complementary visual context from the RGB branch to the motion branch during training, enriching motion representations through training-time visual knowledge transfer (Section~\ref{sec:method:auxiliary-vision}).

\subsection{Modality Representations}
\label{sec:method:modality-representations}

We encode each modality into a sequence of token-level embeddings in a shared latent space. This tokenized representation allows the model to support not only sentence-level retrieval, but also localized motion-language alignment.

\textit{Temporal Encoding of SMPL-X Motion.} 
For the sign motion input, we use a task-relevant subset of SMPL-X \cite{pavlakos2019expressive} parameters as the primary representation. Specifically, following prior SMPL-X-based sign language generation work ~\cite{zuo2025signs, zuo2024simple}, we retain the articulatory components most relevant to sign language, including upper-body, hand, and a low-dimensional facial expression code, while discarding parameters that are less informative for signing. The SMPL-X annotations are provided by ~\cite{zuo2025signs} through a multi stage estimation pipeline, which first estimates rough 3D body poses with OSX \cite{lin2023one} and 3D hand poses with WiLoR \cite{potamias2025wilor}, then refines the upper body joint rotations via reprojection to Mediapipe 2D keypoints \cite{lugaresi2019mediapipe}.

Although SMPL-X provides a structured and appearance-free representation of sign motion, its frame-wise parameters lie in an abstract kinematic space and are not directly suitable for semantic alignment with language. In continuous signing, semantic meaning is often expressed through short-term coordinated movements rather than isolated joints or single frames. To make SMPL-X motion more suitable for semantic alignment with language, we introduce a temporal motion encoding stage, termed the \emph{Movement Encoder}, which aggregates short-term co-articulated dynamics and reduces frame-level redundancy, producing compact motion tokens for subsequent cross-modal alignment.

Given a motion sequence
\begin{equation}
\label{eq:motion-sequence}
M=\{m_1,m_2,\ldots,m_T\}, \quad m_t\in\mathbb{R}^{d_m},
\end{equation}
we first apply the Movement Encoder \(E_m(\cdot)\), implemented as a temporal convolutional encoder, to aggregate short-term coordinated dynamics and reduce the temporal resolution:
\begin{equation}
\label{eq:movement-encoder}
\tilde{M}=E_m(M)\in\mathbb{R}^{\frac{T}{4}\times d_m}.
\end{equation}
The downsampled features are then fed into a Transformer encoder to model longer-range dependencies across motion tokens:
\begin{equation}
\label{eq:motion-transformer}
H_m=\mathrm{Trans}_m(\tilde{M}).
\end{equation}

Finally, the Transformer outputs are mapped into the shared retrieval space:
\begin{equation}
\label{eq:motion-projection}
Z^m=\phi_m(H_m)=\{z^m_1,z^m_2,\ldots,z^m_{N_m}\}, \quad Z^m\in\mathbb{R}^{N_m\times d},
\end{equation}
where \(N_m=T/4\) in our implementation.

\textit{Text representation.}
Given a sentence \(x = \{w_i\}_{i=1}^{N_t}\), we use a pretrained multilingual text encoder, LaBSE~\cite{feng2022language}, to extract contextualized token embeddings:
\begin{equation}
\label{eq:text-encoder}
E_t(x) = \{e^t_1, e^t_2, \dots, e^t_{N_t}\}.
\end{equation}
We retain token-level representations rather than collapsing the sentence into a single global embedding, for two reasons. First, our framework relies on fine-grained motion-language alignment, where localized motion spans need to be associated with semantically relevant textual units. Second, token-level features preserve the compositional structure of the sentence and allow the model to emphasize informative lexical content during cross-modal alignment. The token features are then projected into the shared retrieval space by a lightweight projection head:
\begin{equation}
\label{eq:text-projection}
Z^t = \phi_t(E_t(x)) = \{z^t_1, z^t_2, \dots, z^t_{N_t}\}.
\end{equation}

\textit{RGB representation.}
To provide complementary visual context during training, we introduce an auxiliary RGB branch. Given the RGB video input \(\mathbf{V}\), we first extract frame level visual features using an I3D feature encoder \cite{carreira2017quo}. The resulting features are then fed into a visual encoder to model temporal context, and the outputs are projected into the shared retrieval space:
\begin{equation}
\label{eq:rgb-projection}
\mathbf{Z}^{i} = \phi_i\!\left(\mathrm{Trans}_i(E_i(\mathbf{V}))\right)
= \{\mathbf{z}^{i}_1,\mathbf{z}^{i}_2,\dots,\mathbf{z}^{i}_{N_i}\},
\end{equation}
where \(E_i(\cdot)\) denotes the I3D feature encoder, \(\mathrm{Trans}_i(\cdot)\) denotes the visual encoder, and \(\phi_i(\cdot)\) is the corresponding projection head. Used only during training, this branch provides auxiliary semantic supervision for motion and visual features for distillation.

\textit{Gloss representation.}
For each gloss label \(g_k\), we also construct a textual gloss representation using the same pretrained language encoder followed by a separate projection head:
\begin{equation}
\label{eq:gloss-projection}
\mathbf{Z}_k^g=\phi_g(E_t(g_k)).
\end{equation}
Collecting all gloss labels in a sample yields
\begin{equation}
\label{eq:gloss-collection}
\mathbf{Z}^{g}=\{\mathbf{z}^{g}_1,\mathbf{z}^{g}_2,\dots,\mathbf{z}^{g}_K\}.
\end{equation}
Unlike the sentence text branch, gloss labels are not used as retrieval queries. Instead, they serve as localized semantic anchors for fine-grained alignment with motion spans.

\subsection{Gloss-Guided Boundary-Aware Local Alignment}
\label{sec:method:gloss-guided}

A central challenge in continuous sign language retrieval is that sentence level matching alone cannot capture the internal semantic structure of signing. To reduce semantic interference caused by co articulation, we use gloss temporal spans provided by prior recognition work \cite{chen2022two} to construct local motion language correspondences. Spoken2Sign~\cite{chen2022two} automatically estimates these training-time spans.

\begin{figure*}
  \centering
  \includegraphics[width=0.85\linewidth]{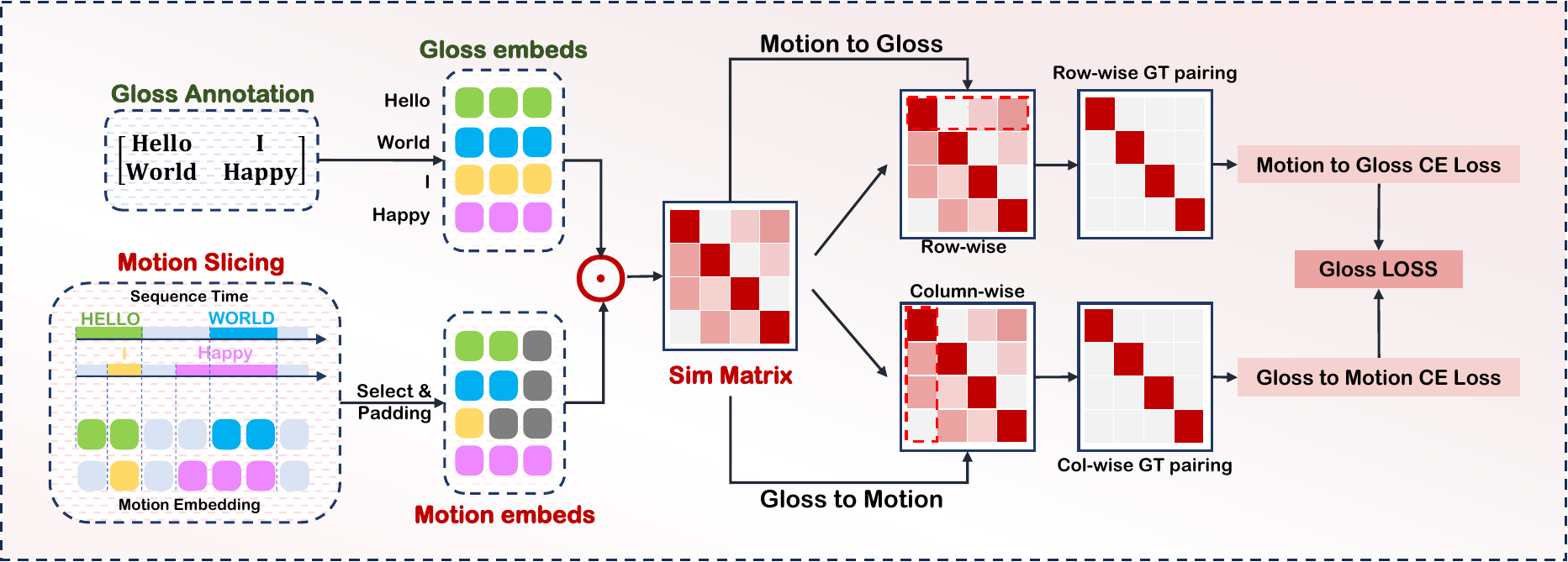}
  \caption{Illustration of the gloss-guided local alignment loss. Motion-token spans are sliced by gloss boundaries and padded for batching. Similarities between gloss and motion-span embeddings form a matrix, where row-wise and column-wise InfoNCE losses enforce motion-to-gloss and gloss-to-motion alignment. Their sum yields the gloss-guided loss $\mathcal{L}_g$.}
  \label{fig:gloss}
\end{figure*}

\textit{Gloss span extraction.}
Using the gloss spans in \(\mathcal{G}\), we map each gloss boundary from the original frame resolution to the motion-token resolution. Let \(u\) denote the temporal downsampling factor (\(u=4\) in our implementation). This yields token-level boundaries:
\begin{equation}
\label{eq:token-boundaries}
\hat{s}_k = \left\lfloor \frac{s_k}{u} \right\rfloor, \qquad
\hat{e}_k = \left\lceil \frac{e_k}{u} \right\rceil.
\end{equation}
We then extract the corresponding motion-token subsequence:
\begin{equation}
\label{eq:motion-span}
\mathbf{Z}^{m}_{k}
=
\{\mathbf{z}^{m}_{t}\mid \hat{s}_k \le t \le \hat{e}_k\}.
\end{equation}

\textit{Bidirectional gloss-motion alignment.}
For each gloss label \(g_k\), we obtain its gloss embedding sequence from the gloss branch, denoted by \(\mathbf{Z}^{g}_{k}\). We then compute a pairwise similarity matrix
\begin{equation}
\label{eq:span-gloss-similarity}
S_{kj} = \mathrm{sim}(\mathbf{Z}^{m}_{k}, \mathbf{Z}^{g}_{j}),
\end{equation}
where $\mathrm{sim}$ is bidirectional late interaction over valid tokens. For valid token sequences $A$ and $B$,
\begin{equation}
\label{eq:late-interaction}
\mathrm{sim}(A,B)=\frac{1}{2}\!\left(
\frac{1}{|A|}\sum_{a\in A}\max_{b\in B}\cos(a,b)
+\frac{1}{|B|}\sum_{b\in B}\max_{a\in A}\cos(b,a)
\right).
\end{equation}
Padding tokens are masked, while repeated gloss labels retain separate motion spans.
Each entry measures the similarity between the \(k\)-th motion span and the \(j\)-th gloss embedding. As shown in Fig.~\ref{fig:gloss}, row-wise normalization corresponds to motion-to-gloss InfoNCE \cite{oord2018representation} with contrastive temperature \textcolor{black}{$\tau$}:
\begin{equation}
\label{eq:motion-to-gloss}
\mathcal{L}_{m\rightarrow g}
=
-\frac{1}{K}\sum_{k=1}^{K}
\log
\frac{
\exp(S_{kk}/\tau)
}{
\sum_{j=1}^{K}\exp(S_{kj}/\tau)
},
\end{equation}
and column-wise normalization yields gloss-to-motion InfoNCE:
\begin{equation}
\label{eq:gloss-to-motion}
\mathcal{L}_{g\rightarrow m}
=
-\frac{1}{K}\sum_{k=1}^{K}
\log
\frac{
\exp(S_{kk}/\tau)
}{
\sum_{j=1}^{K}\exp(S_{jk}/\tau)
}.
\end{equation}
The final gloss-guided local alignment loss is
\begin{equation}
\label{eq:gloss-loss}
\mathcal{L}_{g}=\mathcal{L}_{m\rightarrow g}+\mathcal{L}_{g\rightarrow m}.
\end{equation}

Compared with soft token-frame interactions, this design uses explicit gloss spans as boundary-grounded supervision, yielding cleaner local motion-language alignment under co-articulation.

\subsection{Auxiliary Visual Distillation}
\label{sec:method:auxiliary-vision}

Although SMPL-X provides a structured motion representation, it is more compact than RGB observations and may lose contextual cues. To compensate for this information gap, we introduce an auxiliary visual distillation strategy that uses RGB features only during training \cite{hinton2015distilling, vapnik2015learning, lambert2018deep}. For all cross-modal objectives in this section, we use the same bidirectional InfoNCE loss.

\textit{RGB-text semantic supervision.}
The auxiliary RGB branch is encouraged to align with the sentence text in the shared space:
\begin{equation}
\label{eq:rgb-text-loss}
\mathcal{L}_{it} = \mathrm{InfoNCE}(\mathbf{Z}^{i}, \mathbf{Z}^{t}).
\end{equation}
This term allows the RGB branch to capture sentence-relevant visual semantics.

\textit{Motion-RGB distillation.}
To transfer visual knowledge from RGB to motion, we further align the motion tokens with the RGB tokens:
\begin{equation}
\label{eq:motion-rgb-loss}
\mathcal{L}_{mi} = \mathrm{InfoNCE}(\mathbf{Z}^{m}, \mathbf{Z}^{i}).
\end{equation}
This encourages the motion branch to absorb complementary visual context from RGB while preserving SMPL-X as the primary representation.

\textit{Motion-text retrieval objective.}
The main retrieval objective directly aligns motion tokens with sentence tokens:
\begin{equation}
\label{eq:motion-text-loss}
\mathcal{L}_{mt} = \mathrm{InfoNCE}(\mathbf{Z}^{m}, \mathbf{Z}^{t}),
\end{equation}
which serves as the core supervision for motion-text retrieval.

The overall training objective combines the main motion-text retrieval loss, the auxiliary RGB-text alignment loss, the motion-RGB distillation loss, and the gloss-guided local alignment loss:
\begin{equation}
\label{eq:total-loss}
\mathcal{L}
=
\lambda_{mt}\mathcal{L}_{mt}
+
\lambda_{it}\mathcal{L}_{it}
+
\lambda_{mi}\mathcal{L}_{mi}
+
\lambda_{g}\mathcal{L}_{g},
\end{equation}
where \(\lambda_{mt}\), \(\lambda_{it}\), \(\lambda_{mi}\), and \(\lambda_{g}\) are balancing weights, and the balancing weights are empirically set to 1. 
\section{Experiments}
We evaluate our method on two sign language benchmarks, CSL-Daily and PHOENIX-2014T. We first describe the evaluation protocol and implementation details, then compare with state-of-the-art methods, analyze cross-signer robustness, and present ablation studies on key design choices.
\subsection{Experimental setup}

\textit{Datasets.} 
Following prior sign language retrieval studies~\citep{jiang2024seds,wu2024uncertainty}, we evaluate on CSL-Daily~\citep{zhou2021improving} and PHOENIX-2014T~\citep{camgoz2018neural}, both with gloss annotations. CSL-Daily is a Chinese Sign Language dataset for daily scenarios, containing 18,401/1,077/1,176 training/validation/test samples. It includes multiple videos with the same meaning signed by different signers. PHOENIX-2014T is a German Sign Language dataset of weather forecasts, with 7,096/519/642 samples for training/validation/test.

\textit{Tasks and Metrics.}
Following prior work~\citep{cheng2023cico, jiang2024seds, wu2024uncertainty}, we conduct text-to-video (T2V) and video-to-text (V2T) retrieval. T2V retrieves the matching sign video for a text query, while V2T retrieves the corresponding text for a sign video. Although our method uses SMPL-X as the primary retrieval representation rather than RGB video, we retain the conventional T2V/V2T nomenclature for consistency with existing literature. Performance is measured by R@K ($\uparrow$), which indicates whether the correct paired sample or another semantically equivalent sample is ranked in the top K.

\textit{Implementation Details.}
We implement our model in PyTorch and train it using the AdamW optimizer. The batch size is set to 128, the initial learning rate is \(1 \times 10^{-4}\), and the model is trained for 200 epochs. The learning rate is decayed using CosineAnnealingLR  (\(T_{\max}=400\), \(\eta_{\min}=1 \times 10^{-6}\)), with a weight decay of 0.01. For the text branch, input sentences are tokenized and encoded using LaBSE token embeddings (projected by a token projection head), with a maximum text length of 64. For the motion branch, each sample is represented as an SMPL-X parameter sequence of length 300 and dimension 133, where temporal sequences longer than this limit are uniformly sampled and shorter ones are padded. The shared embedding dimension is set to 512, and the temperature parameter in the contrastive objective is set to 0.07. For prior approaches \cite{cheng2023cico, jiang2024seds, wu2024uncertainty}, we follow their original implementations, performing retrieval in the multimodal RGB video space. In contrast, the proposed method operates on SMPL-X motion representations. 

\subsection{Main comparison}
\textit{Baselines.} We compare our approach with representative state-of-the-art methods for sign language multimodal retrieval, covering global and fine-grained alignment, multimodal fusion, and probabilistic modeling. SPOT-ALIGN~\citep{duarte2022sign} formulates the task as video-text retrieval via global cross-modal embeddings, using iterative sign spotting and representation learning to mitigate limited supervision. CiCo~\citep{cheng2023cico} further models sign-word correspondences and treats retrieval as joint cross-lingual and cross-modal alignment. SEDS~\citep{jiang2024seds} adopts a dual-stream architecture combining RGB and pose to capture global motion and fine-grained articulations. UPRet~\citep{wu2024uncertainty} models cross-modal relationships as probability distributions to account for ambiguity and annotation sparsity. CMCM~\citep{yang2026causality} introduces causal reasoning with multi-grained feature learning and intervention-based strategies to reduce spurious correlations.

\textit{Quantitative results.}
As shown in Table~\ref{tab:retrieval_results_cn}, our method achieves state-of-the-art bidirectional retrieval performance on CSL-Daily, consistently outperforming all previous approaches across both T2V and V2T tasks and all R@K metrics. Importantly, these gains are obtained with SMPL-X motion as the primary retrieval representation and with visual knowledge transferred from the auxiliary RGB branch during training, which directly supports our central claim that reformulating sign language--text alignment in a structured kinematic space can yield stronger semantic matching than appearance-entangled visual representations. The consistent improvements in both retrieval directions suggest that structured human motion provides a more reliable shared embedding space for sign language retrieval.
SiCLIP, the strongest SMPL-X-based baseline reported in MaDiS~\citep{zuo2026madis}, obtains sign-to-text R@1 of 59.10/68.85 on CSL-Daily/PHOENIX-2014T (Supplementary Sec.~B.1).

On the more challenging PHOENIX-2014T dataset, our method remains competitive but trails the strongest RGB-based baselines. We attribute this gap to the lower reliability of SMPL-X motion recovered from broadcast-style videos, where poorer visual quality and motion blur introduce noise into the kinematic representation \cite{moryossef2021evaluating, zimmer2023imposing}. This suggests that kinematics-centric retrieval depends on the quality of motion recovery. Overall, results across the two benchmarks support the promise of SMPL-X as a structured representation for sign language retrieval, especially when accurate 3D motion estimation is available.
Supporting this interpretation, PHOENIX-2014T has higher median acceleration jitter than CSL-Daily (2.43 vs.\ 1.47) (Supplementary Sec.~B.2).
R@1 decreases by at most 1.25 points under SMPL-X parameter noise and 3.42 points under gloss-span perturbations. See Supplementary Sec.~B.3.

\begin{figure*}
  \centering
  \includegraphics[width=\linewidth]{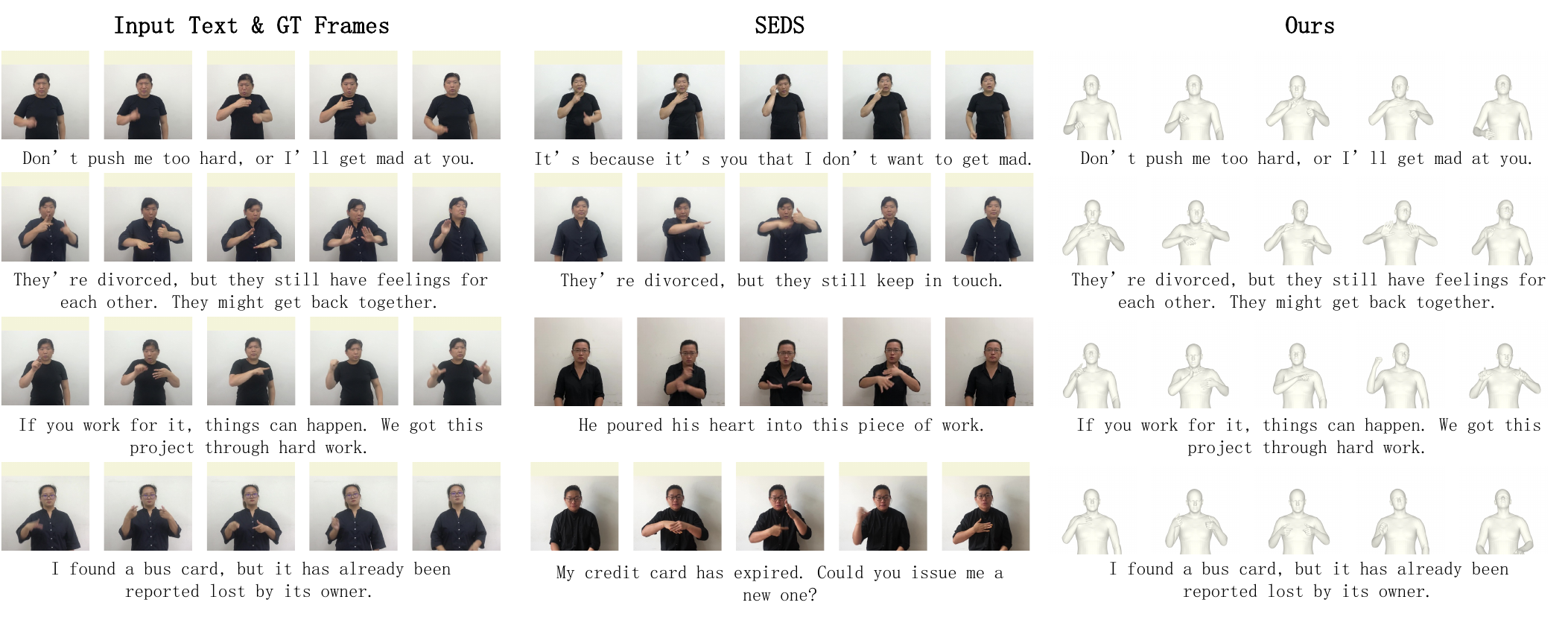}
    \caption{Representative Cases. Given an input text query, we show the GT frames of the matched sample (left), the video retrieved by SEDS with its paired text (middle), and the top-1 SMPL-X retrieved by our method with its paired text (right). 
    }
    \label{fig:compare}
  
\end{figure*}

\textit{Qualitative results.}
Figure~\ref{fig:compare} shows representative cases where SEDS \cite{jiang2024seds} retrieves an incorrect sample while our method retrieves the correct one.
For each text query, we show the ground-truth video frames of the matched sample, the top-1 video retrieved by SEDS with its paired text, and the top-1 SMPL-X motion retrieved by our method with its paired text.
In these examples, the SEDS result is often semantically related to the query but does not match the target sentence.
By contrast, our method retrieves the correct sample, demonstrating the value of structured SMPL-X for fine-grained text-to-sign retrieval.

\begin{table}[t]
\centering
\caption{Comparison with prior methods on CSL-Daily and PHOENIX-2014T (R@K, \%, $\uparrow$). Best results are in bold.}
\label{tab:retrieval_results_cn}
\resizebox{\linewidth}{!}{
\begin{tabular}{llcccccc}
\toprule
\multirow{2}{*}{DataSet} & \multirow{2}{*}{Methods} 
& \multicolumn{3}{c}{T2V} 
& \multicolumn{3}{c}{V2T} \\
\cmidrule(lr){3-5} \cmidrule(lr){6-8}
& & R@1 & R@5 & R@10 & R@1 & R@5 & R@10 \\
\midrule
\multirow{6}{*}{CSL-Daily}
& UPRet \cite{wu2024uncertainty}               & 78.4  & 89.1  & 92.0  & 77.0  & 89.2  & 92.7  \\
& CiCo \cite{cheng2023cico}               & 56.6  & 69.9  & 74.7  & 51.6  & 64.8  & 70.1  \\
& SEDS \cite{jiang2024seds}               & 85.8  & 94.4  & 95.6  & 85.4  & 93.8  & 95.8  \\
& Spot-ALIGN  \cite{duarte2022sign}         & 34.2  & 48.0  & 52.6  & 23.6  & 47.0  & 53.0  \\
& CMCM \cite{yang2026causality}                & 84.7  & 93.3  & 95.1  & 79.3  & 90.2  & 92.7  \\
& Ours                & \textbf{87.1} & \textbf{95.4} & \textbf{96.9} & \textbf{87.0} & \textbf{94.6} & \textbf{96.9} \\
\midrule
\multirow{6}{*}{Phoenix}
& UPRet \cite{wu2024uncertainty}               & 72.0  & 89.1  & 94.1  & 72.0  & 89.4  & 93.3  \\
& CiCo \cite{cheng2023cico}                & 69.5  & 86.6  & 92.1  & 70.2  & 88.0  & 92.8  \\
& SEDS \cite{jiang2024seds}                & \textbf{76.8} & \textbf{91.7} & \textbf{95.3} & \textbf{78.7} & \textbf{92.5} & \textbf{95.2} \\
& Spot-ALIGN \cite{duarte2022sign}         & 55.8  & 79.6  & 87.2  & 53.1  & 79.4  & 86.1  \\
& CMCM \cite{yang2026causality}               & 76.1  & 88.9  & 93.3  & 74.5  & 90.2  & 95.1  \\
& Ours                & 74.1 & 88.1 & 90.6 & 71.2 & 88.3 & 92.1 \\
\bottomrule
\end{tabular}
}
\end{table}

\subsection{Signer-Invariant Representation Analysis}

To analyze whether the kinematic motion representation reduces reliance on signer specific nuisance factors, we study the sign side embedding space on the CSL-Daily test split. PHOENIX-2014T is excluded because it does not contain the same sentence performed by multiple signers. For the RGB based baseline, we extract sample level video embeddings from the RGB branch of SEDS, while for our method we extract sample level motion embeddings from the SMPL-X branch. The analysis is performed on these unimodal embeddings on the sign side, rather than on similarities between text and sign.

Table~\ref{tab:embedding_analysis} summarizes the quantitative analysis. Specifically, \textit{Same-Sentence Diff-Signer} and \textit{Diff-Sentence Same-Signer} respectively reflect cross-signer semantic consistency and signer-specific bias, while \textit{Diff-Sentence Diff-Signer} serves as a general negative baseline. 
As shown in Table~\ref{tab:embedding_analysis}, SMPL-X achieves substantially higher Same-Sentence Diff-Signer similarity while producing much lower Diff-Sentence Same-Signer similarity than RGB. 

\begin{table}[t]
\centering
\caption{Signer-invariance analysis on CSL-Daily. Higher is better for Same-Sentence Diff-Signer; lower is better otherwise.}
\label{tab:embedding_analysis}
\resizebox{0.75\linewidth}{!}{%

\begin{tabular}{lccc}
\toprule
Model & 
\begin{tabular}[c]{@{}c@{}}Same-Sentence\\ Diff-Signer$\uparrow$\end{tabular} &
\begin{tabular}[c]{@{}c@{}}Diff-Sentence\\ Same-Signer$\downarrow$\end{tabular} &
\begin{tabular}[c]{@{}c@{}}Diff-Sentence\\ Diff-Signer$\downarrow$\end{tabular} \\
\midrule
RGB    & 0.7798 & 0.4137 & 0.3651 \\
SMPL-X & \textbf{0.8975} & \textbf{0.0448} & \textbf{0.0041} \\
\bottomrule
\end{tabular}
}
\end{table}

\begin{figure}[t]
  \centering
  \includegraphics[width=1\linewidth]{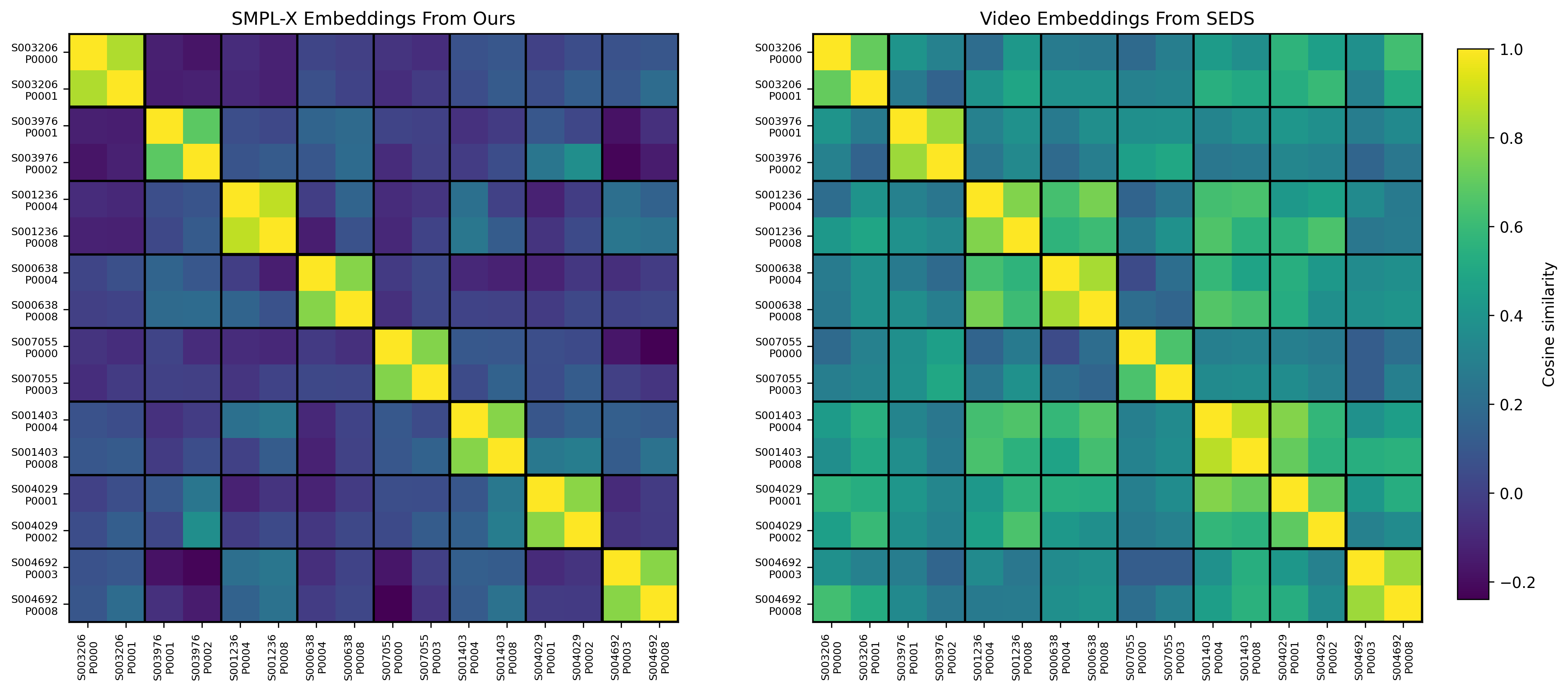}
    \caption{Intra-modal cosine similarity matrices for our SMPL-X embeddings (left) and SEDS RGB embeddings (right) on CSL-Daily. Samples are ordered by sentence identity, forming $2\times2$ diagonal blocks. SMPL-X exhibits clearer diagonal blocks and lower off-diagonal similarities, indicating stronger signer invariance.}
\label{fig:test_heatmap}
\end{figure}

Figure~\ref{fig:test_heatmap} visualizes the intra-modal cosine similarity matrices of sample embeddings from the RGB branch of SEDS and the SMPL-X branch of our model on the same test samples. Samples are ordered by sentence identity, so the diagonal blocks reflect local grouping by sentence semantics. Compared with RGB, the SMPL-X embeddings exhibit clearer diagonal blocks and lower off-block similarities, indicating stronger cross-signer consistency within the same sentence and better separation between different sentence semantics. Together with the quantitative results in Table~\ref{tab:embedding_analysis}, this provides qualitative evidence that SMPL-X is less sensitive to signer-specific factors and yields a more signer-invariant representation space.

\subsection{Ablation Studies}

Table~\ref{tab:ablation} shows that all three components contribute to retrieval. Removing gloss-guided alignment causes the largest R@1 drops, from 87.09 to 68.55 for text-to-motion and from 86.99 to 67.68 for motion-to-text, underscoring the importance of boundary-aware local supervision. Removing the Movement Encoder lowers R@1 to 80.58/81.38, confirming the value of temporal motion encoding, whereas removing RGB-I3D yields smaller reductions to 83.71/85.63, consistent with its auxiliary training role. Corresponding PHOENIX-2014T R@1 ablations are reported in Supplementary Sec.~B.4.

\begin{table}
\centering
\caption{Component ablation on CSL-Daily (R@K, \%, $\uparrow$).}
\resizebox{0.75\linewidth}{!}{%

\begin{tabular}{lcccccc}
\toprule
\multirow{2}{*}{Setting} & \multicolumn{3}{c}{Text to Motion} & \multicolumn{3}{c}{Motion to Text} \\
\cmidrule(lr){2-4} \cmidrule(lr){5-7}
 & R@1 & R@5 & R@10 & R@1 & R@5 & R@10 \\
\midrule
Ours      & 87.09 & 95.36 & 96.87 & 86.99 & 94.64 & 96.85 \\
No MOVE   & 80.58 & 93.86 & 95.86 & 81.38 & 93.20 & 95.92 \\
No I3D    & 83.71 & 95.24 & 96.74 & 85.63 & 94.81 & 96.77 \\
No Gloss  & 68.55 & 86.47 & 90.23 & 67.68 & 85.29 & 90.48 \\
\bottomrule
\end{tabular}
}
\label{tab:ablation}
\end{table}

\section{Conclusion}

In this paper, we presented a gloss-guided kinematic framework for sign language--text alignment that moves beyond appearance-dependent RGB matching and reformulates the task in a structured motion space. By adopting SMPL-X as the primary representation, our method captures the articulated nature of signing in a more semantically meaningful form compatible with motion-centric deployment. To improve fine-grained grounding, we further introduced gloss-guided local alignment, which leverages gloss spans to establish more precise correspondences between localized motion segments and language.

In addition, we developed an auxiliary visual distillation strategy that exploits RGB as privileged supervision only during training, while using the distilled SMPL-X representation for retrieval. This asymmetric design enriches motion representations with complementary visual context through RGB-to-motion distillation. Extensive experiments on CSL-Daily and PHOENIX-2014T demonstrate that the proposed method achieves state-of-the-art bidirectional retrieval performance on CSL-Daily and competitive results on PHOENIX-2014T, highlighting the effectiveness of structured kinematic representations for sign language--text alignment.

In summary, our results suggest that kinematics-centric retrieval is a promising direction for sign language understanding, especially when reliable motion recovery is available. In future work, we plan to further improve the robustness of motion representations under challenging visual conditions and explore broader motion-centric downstream applications, such as sign language generation, avatar animation, and robot learning.
\begin{acks}
We acknowledge ISCRA for awarding this project access to the LEONARDO
supercomputer, owned by the EuroHPC Joint Undertaking, hosted by CINECA
(Italy).
This work was supported by the EU Horizon projects ELIAS (No. 101120237) and ELLIOT (No. 101214398) and by the FIS project GUIDANCE (No. FIS2023-03251).
\end{acks}

\bibliographystyle{ACM-Reference-Format}
\bibliography{sample-base}


\end{document}